# Learning Low Dimensional Convolutional Neural Networks for High-Resolution Remote Sensing Image Retrieval

**Weixun Zhou [1], Shawn Newsam [2], Congmin Li [1], Zhenfeng Shao [1,*]**

[1] State Key Laboratory of Information Engineering in Surveying, Mapping and Remote Sensing, Wuhan University, Wuhan, China
[2] Electrical Engineering and Computer Science, University of California, Merced, CA, USA
e-mails : weixunzhou1990@whu.edu.cn (W.Z.) ; snewsam@ucmerced.edu (S.N.) ; cminlee@whu.edu.cn (C.L.)
\* Correspondence: shaozhenfeng@whu.edu.cn

**Abstract:** Learning powerful feature representations for image retrieval has always been a challenging task in the field of remote sensing. Traditional methods focus on extracting low-level hand-crafted features which are not only time-consuming but also tend to achieve unsatisfactory performance due to the content complexity of remote sensing images. In this paper, we investigate how to extract deep feature representations based on convolutional neural networks (CNN) for high-resolution remote sensing image retrieval (HRRSIR). To this end, two effective schemes are proposed to generate powerful feature representations for HRRSIR. In the first scheme, the deep features are extracted from the fully-connected and convolutional layers of the pre-trained CNN models, respectively; in the second scheme, we propose a novel CNN architecture based on conventional convolution layers and a three-layer perceptron. The novel CNN model is then trained on a large remote sensing dataset to learn low dimensional features. The two schemes are evaluated on several public and challenging datasets, and the results indicate that the proposed schemes and in particular the novel CNN are able to achieve state-of-the-art performance.

## 1. Introduction

With the rapid development of remote sensing sensors over the past few decades, a considerable volume of high-resolution remote sensing image is now available. The high spatial resolution of the image makes it possible for detailed image interpretation and many remote sensing applications. However, how to efficiently organize and manage the huge volume of remote sensing data has always been a challenge in the remote sensing community.

High-resolution remote sensing image retrieval (HRRSIR), which aims to retrieve and return the interested imageries from a large database, is an effective and indispensable method for the management of the large amount of remote sensing data. An integrated HRRSIR system roughly includes two components, feature extraction and similarity measure, and both play an important role. Feature extraction focuses on the generation of powerful feature representations for the image, while similarity measure focuses on feature matching to measure the similarity between the query image and other images in the database.

The focus of this work is feature extraction component due to the fact that the retrieval performance largely depends on whether the features are representative. Traditional HRRSIR methods are mainly based on low-level feature representations, such as global features including spectral features [1] , shape features [2], and especially texture features[3–5], which can achieve satisfactory performance to some extent. In contrast to these global features, local features are generally extracted from image patches centered at the interesting points, thus having desirable properties such as local property, invariance and robustness. Remote sensing image analysis has benefited a lot from these desirable properties of local features, and many methods have been developed for remote sensing registration and detection tasks [6–8]. In addition to these tasks, local features also showed their advantages for HRRSIR. Yang et al. [9] investigated local invariant features for content-based geographic image retrieval for the first time. Extensive experiments on a publicly available dataset indicated their superiority over global features such as simple statistics, color histogram and homogeneous texture. However, both global and local features mentioned above are essentially low-level features. More importantly, these features are hand-crafted features where it is usually a very challenging task to design a powerful feature representation.

Recently, deep learning methods have dramatically improved the state-of-the-art in speech recognition as well as object recognition and detection [10]. Content-based image retrieval also benefits from the success of deep learning [11]. Inspired by such a great success, some researchers start investigating the application of unsupervised deep learning methods for remote sensing recognition tasks such as scene classification [12] and image retrieval [13]. The unsupervised feature learning framework [13] can learn sparse feature representations from the images directly for HRRSIR. Though the performance of the proposed framework is comparative to that of the state-of-the-art, the improvement is limited, which is mainly because the feature learning framework is based on a shallow auto-encoder network with a single hidden layer making it incapable of generating sufficiently powerful feature representations. Therefore, a deep network is necessary in order to generate powerful feature representations for HRRSIR.

Convolutional neural networks (CNN), which generally consists of convolutional, pooling and fully-connected layers, has already been regarded as the most effective deep learning approach due to its remarkable performance on benchmark dataset such as ImageNet [14]. Nevertheless, a large number of labeled training samples and many techniques are needed in order to train a powerful CNN architecture. In practice, a common strategy for this problem is to transfer deep features from the CNN models pre-trained on ImageNet and then apply them to practical applications such as scene classification [15–17] and image retrieval [18–21]. However, whether the deep feature representations extracted from pre-trained CNN models can be used for HRRSIR still remains an open question. Though CNN has been used for remote sensing image retrieval in [21], they only consider features extracted from the last fully-connected layer, while in this paper we investigate how to extract powerful deep features from both the fully-connected and convolutional layers for HRRSIR. To this end, two effective schemes are proposed to generate powerful feature representations for similarity measure. Concisely speaking, in the first scheme, feature representations are extracted from the pre-trained fully-connected and convolutional layers, while in the second scheme, the pre-trained CNN models are fine-tuned on a large-scale remote sensing dataset and then generalized to the target retrieval datasets.

The main contributions of this paper are summarized as follows:

- We propose two effective schemes to explore how to extract powerful deep features using CNN for HRRSIR. In the first scheme, the pre-trained CNN is regarded as a feature extractor, and in the second scheme, a novel CNN architecture is proposed to learn low dimensional features.
- A comparative performance evaluation of the existing pre-trained CNN models using several remote sensing datasets is conducted. Among the datasets, three new challenging datasets are introduced to overcome performance saturation on the existing benchmark dataset.
- The novel CNN is trained on a large remote sensing dataset and then applied to the other remote sensing datasets. The results show that replacing fully-connected layers with multi-layer perceptron can not only decrease the number of parameters but also achieve remarkable performance with low dimensional features.
- The two schemes achieve state-of-the-art performance and can be served as baseline results for HRRSIR using CNN.

The reminder of this paper is organized as follows. In Section 2, we introduce the overall architecture of CNN. In section 3, we present the two proposed schemes for HRRSIR. Experimental results and analysis are shown in Section 4. Section 5 shows some discussions. In section 6, we draw conclusions for this work.

## 2. Convolutional Neural Networks (CNN)

In this section, we first briefly introduce the typical architecture of CNN and then review the pre-trained CNN models evaluated in our work.

*2.1 The Architecture of CNN*

The main building blocks of a CNN architecture consist of different types of layers including convolutional layers, pooling layers and fully-connected layers. There are generally a certain number of filters (also called kernels, weights) of each convolutional layer which can output the same number of feature maps by sliding the filters through feature maps of the previous layer. The polling layers conduct subsampling operation along the spatial dimension of feature maps to reduce the size of feature maps via max or average pooling. The fully-connected layers usually follow the convolutional and polling layers and constitute the final several layers. Figure 1 shows the typical architecture of a CNN model. Note that generally the element-wise rectified linear units (ReLU), i.e. $f(x) = \max(0, x)$ is applied to feature maps of both convolutional and fully-connected layers to generate non-negative features. It is a commonly used activation function in CNN models due to its better performance than the other activation functions [22,23].

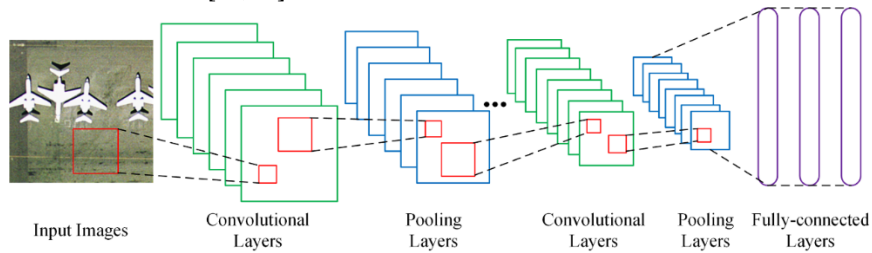

**Figure 1.** The typical architecture of CNN. ReLU layer is ignored here for conciseness.

*2.2 The Pre-trained CNN Models*

Several successful CNN models pre-trained on ImageNet are evaluated in our work, which are the famous baseline model AlexNet [24], the Caffe reference model (CaffeRef) [25], the VGG network [26] and the VGG-VD network[27].

AlexNet was regarded as a baseline model and achieved the best performance in the ImageNet Large Scale Visual Recognition Challenge (ILSVRC-2012). The success of AlexNet is attributed to the large-scale labelled dataset and some techniques such as data augmentation, ReLU activation function and dropout. Dropout is usually used in the first two fully-connected layers to reduce overfitting by randomly setting the output of each hidden neuron to zero with probability 0.5 [28]. AlexNet contains five convolutional layers followed by three fully-connected layers, providing guidance for the design and implementation of successive CNN models.

CaffeRef is trained by the open-source deep learning framework Convolutional Architecture for Fast Feature Embedding (Caffe) [25] with a minor variation from AlexNet. The modifications of CaffeRef lie in the order of pooling and normalization layers as well as data augmentation strategy. It achieved similar performance on ILSVRC-2012 in contrast to AlexNet.

VGG network includes three CNN models, which are VGGF, VGGM and VGGS. These three CNN models are developed to explore a different accuracy/speed trade-off on benchmark datasets for image recognition and object detection. They have similar architectures except for some small differences such as the number and size of filters in some convolutional layers. These models have the same dimensionality of the last hidden layer (the second fully-connected layer) which leads to a 4096-D dimensional feature representation. In order to investigate the effect of feature dimension of the last hidden layer on the final performance, three variant models VGGM128, VGGM1024 and VGGM2048 are trained based on VGGM. In order to speed up training, all the layers except for the second and third layer of VGGM are kept fixed during training. The second fully-connected layers of these three models can generate a 128-D, 1024-D and 2048-D dimensional feature vector, respectively.

Table 1 shows the architectures of these evaluated CNN models for comparison purpose. The readers can refer to relevant papers for more detailed information.

**Table 1**. The architectures of the evaluated CNN Models. Conv1-5 are five convolutional layers and Fc1-3 are three fully-connected layers. For each of the convolutional layers, the first row specifies the number of filters and corresponding filter size as "size×size×number"; the second row indicates the convolution stride and the last row indicates if Local Response Normalization (LRN) is used. For each of the fully-connected layers, its dimensionality is provided. In addition, dropout is applied to Fc1 and Fc2 to overcome overfitting.

| Models | Conv1 | Conv2 | Conv3 | Conv4 | Conv5 | Fc1 | Fc2 | Fc3 |
|---|---|---|---|---|---|---|---|---|
| AlexNet | 11×11×96 stride 4 LRN | 5×5×256 stride 1 LRN | 3×3×384 stride 1 - | 3×3×384 stride 1 - | 3×3×256 stride 1 - | 4096 drop-out | 4096 drop-out | 1000 soft-max |
| CaffeRef | 11×11×96 stride 4 LRN | 5×5×256 stride 1 LRN | 3×3×384 stride 1 - | 3×3×384 stride 1 - | 3×3×256 stride 1 - | 4096 drop-out | 4096 drop-out | 1000 soft-max |
| VGGF | 11×11×64 stride 4 LRN | 5×5×256 stride 1 LRN | 3×3×256 stride 1 - | 3×3×256 stride 1 - | 3×3×256 stride 1 - | 4096 drop-out | 4096 drop-out | 1000 soft-max |
| VGGM | 7×7×96 stride 2 LRN | 5×5×256 stride 2 LRN | 3×3×512 stride 1 - | 3×3×512 stride 1 - | 3×3×512 stride 1 - | 4096 drop-out | 4096 drop-out | 1000 soft-max |
| VGGM-128 | 7×7×96 stride 2 LRN | 5×5×256 stride 2 LRN | 3×3×512 stride 1 - | 3×3×512 stride 1 - | 3×3×512 stride 1 - | 4096 drop-out | 128 drop-out | 1000 soft-max |
| VGGM-1024 | 7×7×96 stride 2 LRN | 5×5×256 stride 2 LRN | 3×3×512 stride 1 - | 3×3×512 stride 1 - | 3×3×512 stride 1 - | 4096 drop-out | 1024 drop-out | 1000 soft-max |
| VGGM-2048 | 7×7×96 stride 2 LRN | 5×5×256 stride 2 LRN | 3×3×512 stride 1 - | 3×3×512 stride 1 - | 3×3×512 stride 1 - | 4096 drop-out | 2048 drop-out | 1000 soft-max |
| VGGS | 7×7×96 stride 2 LRN | 5×5×256 stride 1 - | 3×3×512 stride 1 - | 3×3×512 stride 1 - | 3×3×512 stride 1 - | 4096 drop-out | 4096 drop-out | 1000 soft-max |

VGG-VD network is a very deep CNN network including VD16 (16 weight layers including 13 convolutional layers and 3 fully-connected layers) and VD19 (19 weight layers including 16 convolutional layers and 3 fully-connected layers). These two models are developed to investigate the effect of network depth on its accuracy in the large-scale image recognition setting. It has been demonstrated that the representations extracted by VD16 and VD19 can generalize well to other datasets that are different from ImageNet.

**3. Deep Feature Representations for HRRSIR**

In this section, we present the two proposed schemes for HRRSIR in detail. The package MatConvNet [29] is used for the proposed schemes.

*3.1 First Scheme: Features Extracted from the Pre-trained Layers*

For the first scheme, deep feature representations can be extracted directly from specific layers (fully-connected and convolutional layers) of the pre-trained CNN models. In order to improve performance, some preprocessing steps such as data augmentation and mean subtraction are widely used before feeding the images into CNN. In our work, we just conduct mean subtraction with the mean value provided by corresponding pre-trained CNN.

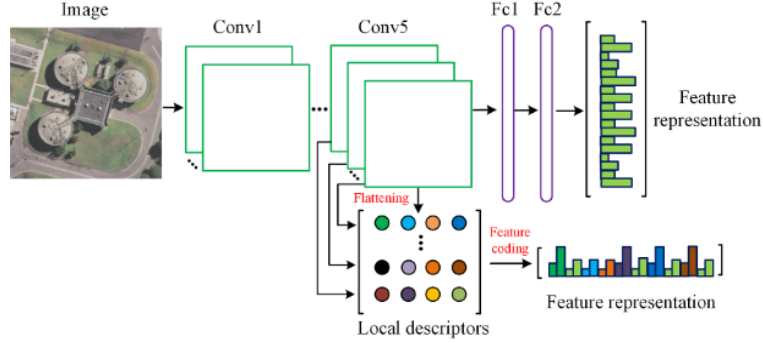

**Figure 2.** Flowchart of the first scheme: deep features extracted from Fc2 and Conv5 layers of the pre-trained CNN model. For conciseness, we refer to features extracted from Fc1-2 and Conv1-5 layers as Fc features (Fc1, Fc2) and Conv features (Conv1, Conv2, Conv3, Conv4, Conv5) respectively.

3.1.1 Features Extracted from Fully-connected Layers

Though there are three fully-connected layers in a pre-trained CNN model, the last layer (Fc3) is usually fed into a softmax (normalized exponential) activation function for classification. Therefore the first two layers (Fc1 and Fc2) are used to extract feature representations in this work, as shown in Figure 2. Both Fc1 and Fc2 can generate a 4096-D dimensional feature representation for all the evaluated models except for the three variant models of VGGM. These 4096-D feature vectors can be directly used for similarity measure to evaluate their retrieval performances.

3.1.2 Features Extracted from Convolutional Layers

Fc features can be considered as global features to some extent, while previous works have demonstrated that local features have better performance than global features when used for HRRSIR[9,30]. Therefore it is of great importance to investigate whether CNN can generate local features and how to aggregate these local descriptors into a compact feature vector. There are some works that investigate how to generate compact features from the activation of fully-connected layers [31] and convolutional layers [32].

Feature maps of the current convolutional layer are computed by sliding the filters through that of the previous layer with a fixed stride, thus each unit of a feature map corresponds to a local region of the image. To achieve the feature representation of this local region, the units of these feature maps need to be recombined. Figure 2 illustrates the process of extracting features from the last convolutional layer (Conv5 layer in this case). The feature maps are firstly flattened to obtain a set of feature vectors. Then each column represents for a local descriptor which can be regarded as the feature representation of the corresponding image region. Let $n$ and $m$ be the number and the size of feature maps respectively, the local descriptors can be defined by:

$$F = [x_1, x_2, ..., x_m] \qquad (1)$$

where $x_i (i=1, 2, ..., m)$ is $n$-dimensional feature vector representing a local descriptor.

The local descriptor set $F$ is of high dimension, thereby using it directly for similarity measure is unpractical. Here, we introduce bag of visual words (BOVW) [33], vector of locally aggregated descriptors (VLAD) [34], improved fisher kernel (IFK) [35] to aggregate these local descriptors into a compact feature vector.

*3.2 Second Scheme: Features Extracted by the Novel Low Dimensional CNN*

In the first scheme CNN is used as a feature extractor to extract Fc and Conv features for HRRSIR, but these existing models are trained on ImageNet which is very different from remote sensing images. In practice, a common strategy for this problem is to fine-tune the pre-trained CNN on the target remote sensing dataset to learn dataset-specific features due to the lack of a large number of labelled images. However, the deep features extracted from the fine-tuned Fc layers are usually 4096-D feature vectors which are not compact enough for large-scale image retrieval and Fc layers are prone to overfitting because most of the parameters lie in Fc layers. In addition, the convolutional filters in CNN are generalized linear models (GLM) based on the assumption that the features are linearly separable, while features that achieve good abstraction are generally highly nonlinear functions of the input.

Network in Network (NIN) [36], the stack of several mlpconv layers, is proposed based on the above assumption. In NIN, the GLM is replaced with an mlpconv layer to enhance model discriminability and the conventional fully-connected layers are replaced with global average pooling to directly output the spatial average of the feature maps from the last mlpconv layer which are then fed into the softmax layer for classification. The readers can refer to [36] for more details.

Inspired by NIN, we propose a novel CNN that has high model discriminability and can generate low dimensional features. Figure 3 shows the overall structure of the low dimensional CNN (LDCNN) which consists of five conventional convolution layers, an mlpconv layer, a global average pooling layer and a softmax classifier layer. LDCNN is essentially the combination of conventional CNN (linear convolution layer) and NIN (mlpconv layer). The structure design is based on the assumption that in conventional CNN the first several layers are trained to learn low-level features such as edges and corners that are linearly separable while the last several layers are trained to learn more abstract high-level features that are nonlinearly separable. The mlpconv layer we used in this paper is a three-layer perceptron trainable by back-propagation and is able to generate one feature map for each corresponding class. The global average pooling layer will take the average of each feature map and lead to an $n$-dimensional feature vector ($n$ is the number of image classes) which will be used for HRRSIR in this paper.

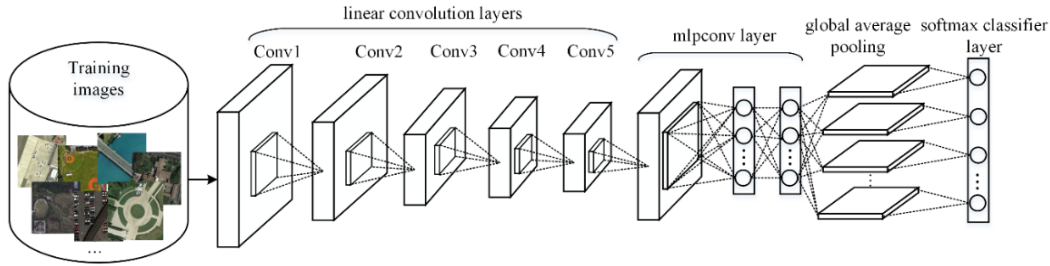

**Figure 3.** The overall structure of the proposed CNN architecture. There are five linear convolution layers and an mlpconv layer followed by a global average pooling layer.

## 4. Experiments and Analysis

In this section, we evaluate the performance of the proposed schemes for HRRSIR on several publicly available remote sensing image datasets. We first introduce the datasets and experimental setup and then present the experimental results in detail.

*4.1 Dataset*

UC Merced dataset (UCMD) [9] is a challenging dataset containing 21 image classes, which are agricultural, airplane, baseball diamond, beach, buildings, chaparral, dense residential, forest, freeway, golf course, harbor, intersection, medium density residential, mobile home park, overpass, parking lot, river, runway, sparse residential, storage tanks, and tennis courts. Each class has 100 images with the size of 256 × 256 pixels and the resolution of about one foot. This dataset is cropped from large aerial images downloaded from USGS. Figure 4 shows some example images of this dataset.

WHU-RS dataset (RSD) [37] is collected from Google Earth (Google Inc.), which consists of 19 classes: airport, beach, bridge, commercial area, desert, farmland, football field, forest, industrial area, meadow, mountain, park, parking, pond, port, railway station, residential area, river and viaduct. There are a total of 1005 images with the size of 600*600 pixels. Figure 5 shows some example images of this dataset.

RSSCN7 dataset [38] consists of 7 land-use classes, which are grassland, forest, farmland, parking lot, residential region, industrial region as well as river and lake. For each image class, there are 400 images with the size of 400×400 pixels sampled on four different zoom scales with 100 images per scale. Figure 6 shows some example images of this dataset.

AID dataset [39] is a large-scale publicly available dataset in contrast to the three datasets mentioned above. It is collected for advancing the state-of-the-arts in scene classification of remote sensing images. This dataset consists of 30 scene types: airport, bare land, baseball field, beach, bridge, center, church, commercial, dense residential, desert, farmland, forest, industrial, meadow, medium residential, mountain, park, parking, playground, pond, port, railway station, resort, river, school, sparse residential, square, stadium, storage tanks and viaduct. There are a total of 10000 images and each class has 220 to 420 images with the size of 600×600 pixels. This dataset is very challenging because of its scale and multi-resolution. Figure 7 shows some example images of AID.

UCMD is a widely used dataset for retrieval performance evaluation, but it is a relatively small dataset and the results on this dataset have been saturated. In contrast to UCMD, the other three datasets are more challenging due to the image scale, image size and pixel resolution, thus they are suitable for performance evaluation of deep features from various CNN models.

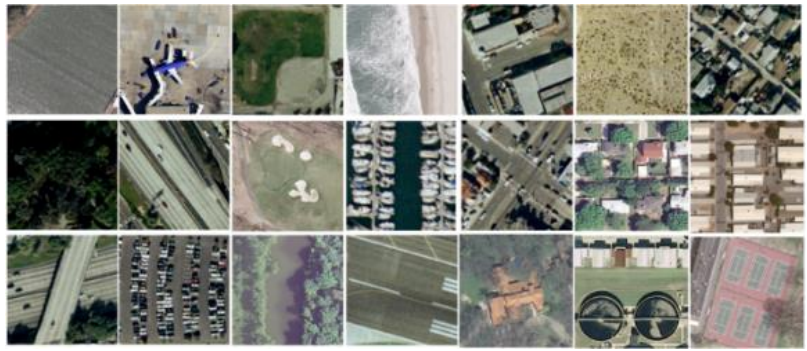

**Figure 4**. Some examples of UCMD. From the top left to bottom right: agricultural, airplane, baseball diamond, beach, buildings, chaparral, dense residential, forest, freeway, golf course, harbor, intersection, medium density residential, mobile home park, overpass, parking lot, river, runway, sparse residential, storage tanks, and tennis courts.

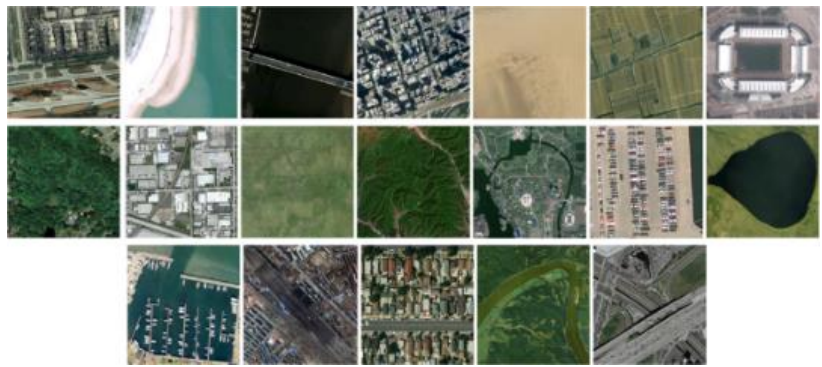

**Figure 5**. Some examples of RSD. From the top left to bottom right: airport, beach, bridge, commercial area, desert, farmland, football field, forest, industrial area, meadow, mountain, park, parking, pond, port, railway station, residential area, river and viaduct.

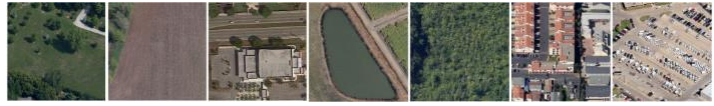

**Figure 6**. Some examples of RSSCN7. From left to right: grass, field, industry, lake, resident, parking.

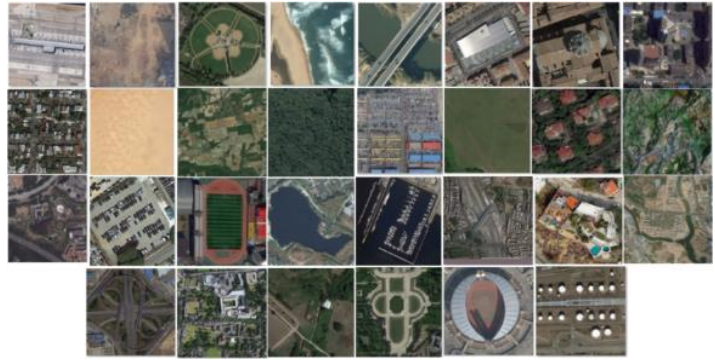

**Figure 7**. Some examples from AID. From the top left to bottom right: airport, bare land, baseball field, beach, bridge, center, church, commercial, dense residential, desert, farmland, forest, industrial, meadow, medium residential, mountain, park, parking, playground, pond, port, railway station, resort, river, school, sparse residential, square, stadium, storage tanks and viaduct.

*4.2 Experimental Setup*

*4.2.1 Implementation Details*

In the first scheme, the images are resized to 227×227 pixels for AlexNet and CaffeRef and to 224×224 pixels for the other networks due to the required input dimension of these CNN models. K-means clustering is used to construct the dictionary for Conv features and the dictionary sizes of BOVW, VLAD and IFK are empirically assigned to be 1000, 100 and 100 respectively.

Regarding the training of LDCNN, the weights of the five convolutional layers are transferred from VGGM. We also tried AlexNet, CaffeRef, VGGF and VGGS but achieve comparative or even worse performance. The weights of convolutional layers are kept fixed in order to speed up training with the limited size of labelled remote sensing images. More specially, the layers after the last pooling layer of VGGM are removed and the rest layers are reserved for LDCNN. The weights of the mlpconv layer are initialized from a Gaussian distribution (the mean is 0 and the standard variance is 0.01). The initial learning rate is set to 0.001 and is lowered by a scale of 10 when the accuracy on the training set stops improving, and dropout is applied to the mlpconv layer. AID dataset is used to train LDCNN because it has 10000 image in total. The training images are randomly selected from AID dataset with 80% images per class and the rest constitute the testing images to indicate when to stop training.

In the following experiments, we perform 2100 queries for UCMD dataset, 1005 queries for RSD dataset and 2800 queries for RSSCN7 dataset while in the case of AID dataset we evaluate 10000 queries in total. The simple metric Euclidean distance is used as the similarity measure, and all the feature vectors are *L*2 normalized before similarity measure.

*4.2.2 Performance Measures*

The average normalized modified retrieval rank (ANMRR) and mean average precision (mAP) are used to evaluate the retrieval performance.

Let $q$ be a query image with the ground truth size of $NG(q)$, $R(k)$ be the retrieved rank of the *k-th* image, which is defined as

$$R(k) = \begin{cases} R(k), & R(k) \leq K(q) \\ 1.25K(q), & R(k) > K(q) \end{cases} \quad (2)$$

where $K(q) = 2NG(q)$ is used to impose a penalty on the retrieved images with a higher rank. The normalized modified retrieval rank (NMRR) is defined as

$$NMRR(q) = \frac{AR(q) - 0.5[1 + NG(q)]}{1.25K(q) - 0.5[1 + NG(q)]} \quad (3)$$

where $AR(q) = \frac{1}{NG(q)} \sum_{k=1}^{NG(q)} R(k)$ is the average rank. Then the final ANMRR can be defined as

$$ANMRR = \frac{1}{NQ} \sum_{q=1}^{NQ} NMRR(q) \quad (4)$$

where $NQ$ is the number of queries. ANMRR ranges from zero to one, and a lower value means better retrieval performance.

Given a set of $Q$ queries, mAP is defined as

$$mAP = \frac{\sum_{q=1}^{Q} AveP(q)}{Q} \quad (5)$$

where $AveP$ is the average precision defined as

$$AveP = \frac{\sum_{k=1}^{n} (P(k) \times rel(k))}{number\ of\ relevant\ images} \quad (6)$$

where $P(k)$ is the precision at cutoff $k$, $rel(k)$ is an indicator function equaling 1 if the image at rank $k$ is a relevant image and zero otherwise. $k$ and $n$ are the rank and number of the retrieved images, respectively. Note that the average is over all relevant images.

We also use precision at $k$ ($P@k$) as an auxiliary performance measure in the case that ANMRR and mAP achieve opposite results. Precision is defined as the fraction of retrieved images that are relevant to the query image.

*4.3 Results of the First Scheme*

4.3.1 Results of Fc Features

The results of Fc features on the four datasets are shown in Table 2. Regarding UCMD dataset, the best result is obtained by using Fc2 feature of VGGM, which achieves an ANMRR value that is about 12% lower and a mAP value that is about 14% higher than that of the worst result achieved by VGGM128_Fc2. Regarding RSD dataset, the best result is obtained by using Fc2 feature of CaffeRef, which achieves an ANMRR value that is about 18% lower and a mAP value that is about 21% higher than that of the worst result achieved by VGGM128_Fc2.

Note that sometimes the two performance measures ANMRR and mAP are likely to achieve "opposite" results. For example, VGGF_Fc2 and VGGM_Fc2 achieve better results on RSSCN7 dataset than the other features in terms of ANMRR value, while VGGM_Fc1 performs the best on RSSCN7 dataset in terms of mAP value. The "opposite" results can also be found on AID dataset in terms of VGGS_Fc1 and VGGS_Fc2 features. Here $P@k$ is used to further investigate the performances of these features, as shown in Table 3. When the number of retrieved images $k$ is within 100, we can see the Fc features of VGGM and in particular VGGM_Fc1 performs slightly better than VGGF_Fc2 in the case of RSSCN7 dataset, and in the case of AID dataset, VGGS_Fc1 performs slightly better than VGGS_Fc2.

It is interesting to find that VGGM performs better than its three variant models on the four datasets, indicating the lower dimension of Fc2 feature does not improve the performance. However, in contrast to VGGM, these three variant models have small storage and query cost due to the lower feature dimension. It can be also observed that Fc2 feature performs better than Fc1 feature for most of the evaluated CNN models on the four datasets.

**Table 2.** The performances of Fc features (ReLU is used) extracted by different CNN models on the four datasets. For ANMRR, the lower is the value the better is the performance, while for mAP, the bigger is the value the better is the performance. The best result is reported in bold.

| Features | UCMD | | RSD | | RSSCN7 | | AID | |
|---|---|---|---|---|---|---|---|---|
| | ANMRR | mAP | ANMRR | mAP | ANMRR | mAP | ANMRR | mAP |
| AlexNet_Fc1 | 0.447 | 0.4783 | 0.332 | 0.5960 | 0.474 | 0.4120 | 0.548 | 0.3532 |
| AlexNet_Fc2 | 0.410 | 0.5113 | 0.304 | 0.6206 | 0.446 | 0.4329 | 0.534 | 0.3614 |
| CaffeRef_Fc1 | 0.429 | 0.4982 | 0.305 | 0.6273 | 0.446 | 0.4381 | 0.532 | 0.3692 |
| CaffeRef_Fc2 | 0.402 | 0.5200 | **0.283** | **0.6460** | 0.433 | 0.4474 | 0.526 | 0.3694 |
| VGGF_Fc1 | 0.417 | 0.5116 | 0.302 | 0.6283 | 0.450 | 0.4346 | 0.534 | 0.3674 |
| VGGF_Fc2 | 0.386 | 0.5355 | 0.288 | 0.6399 | **0.440** | 0.4400 | 0.527 | 0.3694 |
| VGGM_Fc1 | 0.404 | 0.5235 | 0.305 | 0.6241 | 0.442 | **0.4479** | 0.526 | 0.3760 |
| VGGM_Fc2 | **0.378** | **0.5444** | 0.300 | 0.6255 | **0.440** | 0.4412 | 0.533 | 0.3632 |
| VGGM128_Fc1 | 0.435 | 0.4863 | 0.356 | 0.5599 | 0.465 | 0.4176 | 0.582 | 0.3145 |
| VGGM128_Fc2 | 0.498 | 0.4093 | 0.463 | 0.4393 | 0.513 | 0.3606 | 0.676 | 0.2183 |
| VGGM1024_Fc1 | 0.413 | 0.5138 | 0.321 | 0.6052 | 0.454 | 0.4380 | 0.542 | 0.3590 |
| VGGM1024_Fc2 | 0.400 | 0.5165 | 0.330 | 0.5891 | 0.447 | 0.4337 | 0.568 | 0.3249 |
| VGGM2048_Fc1 | 0.414 | 0.5130 | 0.317 | 0.6110 | 0.455 | 0.4365 | 0.536 | 0.3662 |
| VGGM2048_Fc2 | 0.388 | 0.5315 | 0.316 | 0.6053 | 0.446 | 0.4357 | 0.552 | 0.3426 |
| VGGS_Fc1 | 0.410 | 0.5173 | 0.307 | 0.6224 | 0.449 | 0.4406 | 0.526 | **0.3761** |
| VGGS_Fc2 | 0.381 | 0.5417 | 0.296 | 0.6288 | 0.441 | 0.4412 | **0.523** | 0.3725 |
| VD16_Fc1 | 0.399 | 0.5252 | 0.316 | 0.6102 | 0.444 | 0.4354 | 0.548 | 0.3516 |
| VD16_Fc2 | 0.394 | 0.5247 | 0.324 | 0.5974 | 0.452 | 0.4241 | 0.568 | 0.3272 |
| VD19_Fc1 | 0.408 | 0.5144 | 0.336 | 0.5843 | 0.454 | 0.4243 | 0.554 | 0.3457 |
| VD19_Fc2 | 0.398 | 0.5195 | 0.342 | 0.5736 | 0.457 | 0.4173 | 0.570 | 0.3255 |

**Table 3.** The *P@k* values of Fc features that achieve opposite results on RSSCN7 and AID datasets by using ANMRR and mAP measures.

| Measures | RSSCN7 | | | AID | |
|---|---|---|---|---|---|
| | VGGF_Fc2 | VGGM_Fc1 | VGGM_Fc2 | VGGS_Fc1 | VGGS_Fc2 |
| P@5 | 0.7974 | 0.8098 | 0.8007 | 0.7754 | 0.7685 |
| P@10 | 0.7645 | 0.7787 | 0.7687 | 0.7415 | 0.7346 |
| P@50 | 0.6618 | 0.6723 | 0.6626 | 0.6265 | 0.6214 |
| P@100 | 0.5940 | 0.6040 | 0.5943 | 0.5550 | 0.5521 |
| P@1000 | 0.2960 | 0.2915 | 0.2962 | 0.2069 | 0.2097 |

**Table 4**. The performances of Conv features (ReLU is used) aggregated using BOVW, VLAD and IFK on the four datasets. For ANMRR, the lower is the value the better is the performance, while for mAP, the bigger is the value the better is the performance. The best result is reported in bold.

| Features | UCMD | | RSD | | RSSCN7 | | AID | |
|---|---|---|---|---|---|---|---|---|
| | ANMRR | mAP | ANMRR | mAP | ANMRR | mAP | ANMRR | mAP |
| AlexNet_BOVW | 0.594 | 0.3240 | 0.539 | 0.3715 | 0.552 | 0.3403 | 0.699 | 0.2058 |
| AlexNet_VLAD | 0.551 | 0.3538 | 0.419 | 0.4921 | 0.465 | 0.4144 | 0.616 | 0.2793 |
| AlexNet_IFK | 0.500 | 0.4217 | 0.417 | 0.4958 | 0.486 | 0.4007 | 0.642 | 0.2592 |
| CaffeRef_BOVW | 0.571 | 0.3416 | 0.493 | 0.4128 | 0.493 | 0.3858 | 0.675 | 0.2265 |
| CaffeRef_VLAD | 0.563 | 0.3396 | 0.364 | 0.5552 | 0.428 | 0.4543 | 0.591 | 0.3047 |
| CaffeRef_IFK | 0.461 | 0.4595 | 0.364 | 0.5562 | 0.428 | **0.4628** | 0.601 | 0.2998 |
| VGGF_BOVW | 0.554 | 0.3578 | 0.479 | 0.4217 | 0.486 | 0.3926 | 0.682 | 0.2181 |
| VGGF_VLAD | 0.553 | 0.3483 | 0.374 | 0.5408 | 0.444 | 0.4352 | 0.604 | 0.2895 |
| VGGF_IFK | 0.475 | 0.4464 | 0.397 | 0.5141 | 0.454 | 0.4327 | 0.620 | 0.2769 |
| VGGM_BOVW | 0.590 | 0.3237 | 0.531 | 0.3790 | 0.549 | 0.3469 | 0.699 | 0.2054 |
| VGGM_VLAD | 0.531 | 0.3701 | 0.352 | 0.5640 | **0.420** | 0.4621 | 0.572 | 0.3200 |
| VGGM_IFK | 0.458 | 0.4620 | 0.382 | 0.5336 | 0.431 | 0.4516 | 0.605 | 0.2955 |
| VGGM128_BOVW | 0.656 | 0.2717 | 0.606 | 0.3120 | 0.596 | 0.3152 | 0.743 | 0.1728 |
| VGGM128_VLAD | 0.536 | 0.3683 | 0.373 | 0.5377 | 0.436 | 0.4411 | 0.602 | 0.2891 |
| VGGM128_IFK | 0.471 | 0.4437 | 0.424 | 0.4834 | 0.442 | 0.4388 | 0.649 | 0.2514 |
| VGGM1024_BOVW | 0.622 | 0.2987 | 0.558 | 0.3565 | 0.564 | 0.3393 | 0.714 | 0.1941 |
| VGGM1024_VLAD | 0.535 | 0.3691 | 0.358 | 0.5560 | 0.425 | 0.4568 | 0.577 | 0.3160 |
| VGGM1024_IFK | 0.454 | 0.4637 | 0.399 | 0.5149 | 0.436 | 0.4490 | 0.619 | 0.2809 |
| VGGM2048_BOVW | 0.609 | 0.3116 | 0.548 | 0.3713 | 0.566 | 0.3378 | 0.715 | 0.1939 |
| VGGM2048_VLAD | 0.531 | 0.3728 | 0.354 | 0.5591 | 0.423 | 0.4594 | 0.576 | 0.3150 |
| VGGM2048_IFK | 0.456 | 0.4643 | 0.376 | 0.5387 | 0.430 | 0.4535 | 0.610 | 0.2907 |
| VGGS_BOVW | 0.615 | 0.3021 | 0.521 | 0.3843 | 0.522 | 0.3711 | 0.670 | 0.2347 |
| VGGS_VLAD | 0.526 | 0.3763 | 0.355 | 0.5563 | 0.428 | 0.4541 | 0.555 | **0.3396** |
| VGGS_IFK | 0.453 | 0.4666 | 0.368 | 0.5497 | 0.428 | 0.4574 | 0.596 | 0.3054 |
| VD16_BOVW | 0.518 | 0.3909 | 0.500 | 0.3990 | 0.478 | 0.3930 | 0.677 | 0.2236 |
| VD16_VLAD | 0.533 | 0.3666 | **0.342** | **0.5724** | 0.426 | 0.4436 | **0.554** | 0.3365 |
| VD16_IFK | **0.407** | **0.5102** | 0.368 | 0.5478 | 0.436 | 0.4405 | 0.594 | 0.3060 |
| VD19_BOVW | 0.530 | 0.3768 | 0.509 | 0.3896 | 0.498 | 0.3762 | 0.673 | 0.2253 |
| VD19_VLAD | 0.514 | 0.3837 | 0.358 | 0.5533 | 0.439 | 0.4310 | 0.562 | 0.3305 |
| VD19_IFK | 0.423 | 0.4908 | 0.375 | 0.5411 | 0.440 | 0.4375 | 0.604 | 0.2965 |

**Table 5.** The *P@k* values of Conv features that achieve opposite results on RSSCN7 and AID datasets by using ANMRR and mAP measures.

| Measures | RSSCN7 | | AID | |
|---|---|---|---|---|
| | CaffeRef_IFK | VGGM_VLAD | VGGS_VLAD | VD16_VLAD |
| P@5 | 0.7741 | 0.7997 | 0.7013 | 0.6870 |
| P@10 | 0.7477 | 0.7713 | 0.6662 | 0.6577 |
| P@50 | 0.6606 | 0.6886 | 0.5649 | 0.5654 |
| P@100 | 0.6024 | 0.6294 | 0.5049 | 0.5081 |
| P@1000 | 0.2977 | 0.2957 | 0.2004 | 0.1997 |

4.3.2 Results of Conv Features

Table 4 shows the results of Conv features aggregated using BOVW, VLAD and IFK on the four datasets. Regarding UCMD dataset, IFK performs better than BOVW and VLAD for all the evaluated CNN models and the best result is achieved by VD16 (0.407 measured in ANMRR value). It can also be observed that BOVW performs the worst except for VD16. This makes sense because BOVW ignores spatial information which is very important for remote sensing images while encoding local descriptors into a compact feature vector. Regarding the other three datasets, VLAD has better performance than BOVW and IFK for most of the evaluated CNN models and the best results on RSD, RSSCN7 and AID datasets are achieved by VD16 (0.342 measured in ANMRR value), VGGM (0.420 measured in ANMRR value) and VD16 (0.554 measured in ANMRR value), respectively. Moreover, we can see BOVW still has the worst performance among these three feature aggregation methods.

As the performances of Fc features on RSSCN7 and AID datasets, Conv features also achieve some "opposite" results on these two datasets. For example, VGGM_VLAD performs the best on RSSCN7 dataset in terms of ANMRR value, while CaffeRef_IFK outperforms the other features in terms of mAP value. The "opposite" results can also be found on AID dataset with respect to VD16_VLAD and VGGS_VLAD features. Here *P@k* is also used to further investigate the performances of these features, as shown in Table 5. It is clear that VGGM_VLAD achieves better performance than CaffeRef_IFK when the number of retrieved images are within 100. In the case of AID dataset, VGGS_VLAD has better performance when the number of retrieved images are within 10 or beyond 1000.

4.3.3 Effect of ReLU on Fc and Conv Features

Both Fc and Conv features can extract with or without the use of ReLU transformation. Deep features with ReLU mean ReLU is applied to generate non-negative features and improve nonlinearity, while deep features without ReLU mean the opposite. To give an extensive evaluation of the deep features, the effect of ReLU on Fc and Conv features is investigated.

Figure 8 shows the results of Fc features extracted by different CNN models on the four datasets. We can see Fc1 features improve the results on these datasets in contrast to Fc1_ReLU features, indicating the nonlinearity does not contribute to the performance improvement as expected. However, we tend to achieve opposite results (Fc2_ReLU achieves better or similar performance compared with Fc2) in terms of Fc2 and Fc2_ReLU features except for VGGM128, VGGM1024 and VGGM2048. A possible explanation is that Fc2 layer can generate higher-level features which are then used as the input of the classification layer (Fc3 layer) thus improving nonlinearity will benefit the final results.

Figure 9 shows the results of Conv features aggregated by BOVW, VLAD and IFK on the four datasets. It can be observed that the use of ReLU (BOVW_ReLU, VLAD_ReLU and IFK_ReLU) decreases the performance of the evaluated CNN models on the four datasets except for UCMD dataset. Regarding UCMD dataset, the use of ReLU decreases the performance in the case of BOVW but improves the performance in the case of VLAD (except for VD16 and VD19) and IFK. In addition, it is interesting to find that BOVW achieves better performance than VLAD and IFK on UCMD dataset and IFK performs better than BOVW and VLAD on the other three datasets in terms of Conv features without the use of ReLU.

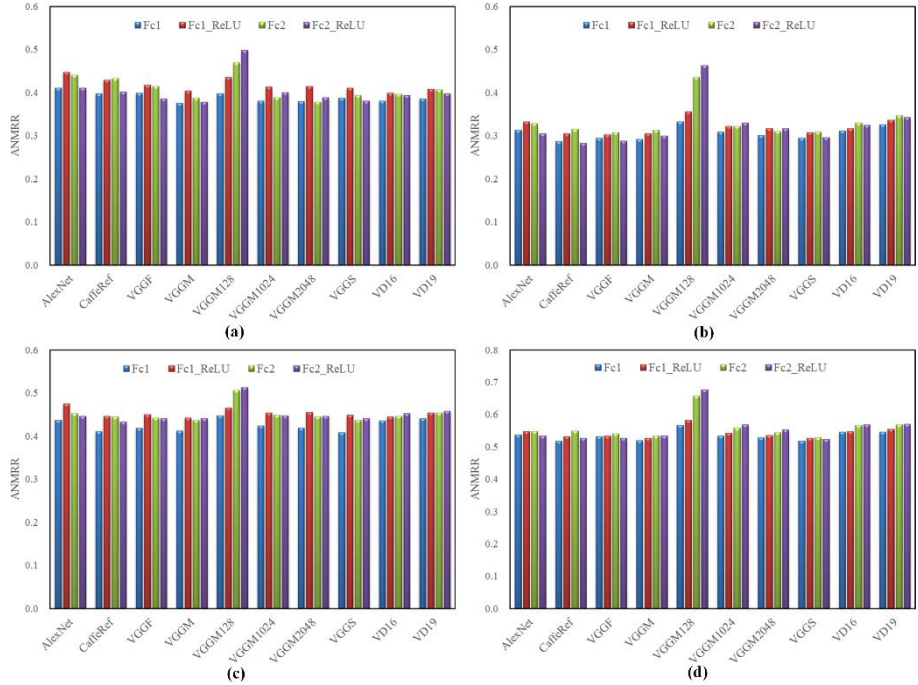

**Figure 8**. The effect of ReLU on Fc1 and Fc2 features of each evaluated CNN. **(a)** Result on UCMD dataset; **(b)** Result on RSD dataset; **(c)** Result on RSSCN7 dataset; **(d)** Result on AID dataset. For Fc1_ReLU and Fc2_ReLU features, ReLU is applied to the extracted Fc features.

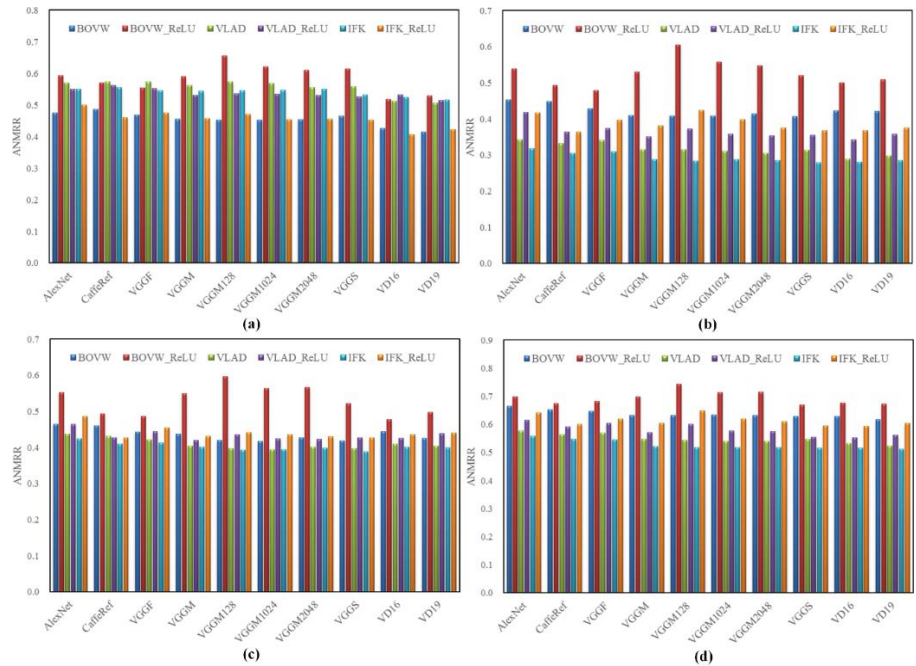

**Figure 9**. The effect of ReLU on Conv features. **(a)** Result on UCMD dataset; **(b)** Result on RSD dataset; **(c)** Result on RSSCN7 dataset; **(d)** Result on AID dataset. For BOVW_ReLU, VLAD_ReLU and IFK_ReLU features, ReLU is applied to Conv features before feature aggregation.

Table 6 shows the best results of Fc and Conv features on the four datasets. The models that achieve these results are also shown in the table. It is interesting to find that the very deep models (VD16 and VD19) perform worse than the relatively shallow models in terms of Fc features but perform better in terms of Conv features. These results indicate that network depth has an effect on the retrieval performance. Specially, increasing the convolutional layer depth will improve the performance of Conv features but will decrease the performance of Fc features. For example, VD16 has 13 convolutional layers and 3 fully-connected layers while VGGS has 5 convolutional layers and 3 fully-connected layers, however, VD16 achieves better performance on AID dataset in terms of Conv features but worse performance in terms of Fc features. This makes sense because Fc layers can generate high-level dataset-specific features while Conv layers can generate more generic features. In addition, more data and techniques are needed in order to train a successful very deep model such as VD16 and VD19 than those relatively shallow models. Moreover, very deep models will also have higher time consumption than those relatively shallow models. Therefore it is wise to balance the tradeoff between the performance and time consumption in practice especially for large-scale tasks such as image recognition and image retrieval.

Table 6. The models that achieve the best results in terms of Fc and Conv features on each dataset. The numbers are ANMRR values.

| Features | UCMD | | RSD | | RSSCN7 | | AID | |
|---|---|---|---|---|---|---|---|---|
| Fc1 | VGGM | 0.375 | Caffe_Ref | 0.286 | VGGS | 0.408 | VGGS | 0.518 |
| Fc1_ReLU | VD16 | 0.399 | VGGF | 0.302 | VGGM | 0.442 | VGGS | 0.526 |
| Fc2 | VGGM2048 | 0.378 | VGGF | 0.307 | VGGS | 0.437 | VGGS | 0.528 |
| Fc2_ReLU | VGGM | 0.378 | Caffe_Ref | 0.283 | Caffe_Ref | 0.433 | VGGS | 0.523 |
| Conv | VD19 | 0.415 | VGGS | 0.279 | VGGS | 0.388 | VD19 | 0.512 |
| Conv_ReLU | VD16 | 0.407 | VD16 | 0.342 | VGGM | 0.420 | VD16 | 0.554 |

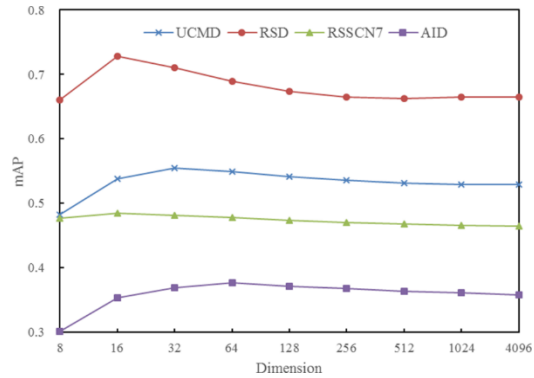

Figure 10. The results of Fc features with various dimensions on the four datasets. The numbers of y axis are mAP values. VGGM_Fc1 is used for UCMD, CaffeRef_Fc2_ReLU is used for RSD and VGGS_Fc1 is used for RSSCN7 and AID.

4.3.4 Results of Low Dimensional Fc Features

The fully-connected layers of these evaluated CNN models result in 4096-D dimensional feature vectors except for the variant models of VGGM. Though these Fc features are already rather compact compared to Conv features, it is necessary to extract more compact features in order to save storage cost and improve retrieval efficiency especially for large-scale image retrieval. To this end, principal component analysis (PCA) is used to extract low dimensional Fc features.

For each dataset, PCA is applied to Fc feature that achieves the best performance as shown in Table 6. Figure 10 shows the results of these Fc features with various dimensions ranging from 8-D to 4096-D. We can see the low dimensional Fc features tend to achieve comparative performance in contrast to the 4096-D features. More specially, the 16-D Fc features perform slightly better than the other Fc features on RSD and RSSCN7 datasets, while the 32-D and 64-D Fc features achieve slightly better performance on UCMD and AID datasets respectively. These results indicate that low dimensional Fc features can not only benefit the retrieval performance but also save the storage cost and improve retrieval efficiency which are important for large-scale image retrieval.

4.3.5 Comparisons with State-of-the-Art Methods

As shown in Table 7, we compare the best result (0.375 achieved by 4096-D Fc1 feature of VGGM) of deep features on UCMD dataset with several state-of-the-art methods which are local invariant features [9], VLAD-PQ [40] and morphological texture [5]. We can see the deep features result in remarkable performance and improve the state-of-the-art by a significant margin.

**Table 7.** Comparisons of deep features with state-of-the-art methods on UCMD dataset. The numbers are ANMRR values.

| Deep features | Local features | VLAD-PQ | Morphological Texture |
|---|---|---|---|
| 0.375 | 0.591 | 0.451 | 0.575 |

**Table 8.** Results of the proposed LDCNN, the first scheme and the fine-tuned VGGM on UCMD, RSD and RSSCN7 datasets. The numbers are ANMRR values. For the first scheme, we choose the best results as shown in Table 6; for VGGM-Finetune, VGGM is fine-tuned on AID dataset.

| Datasets | First Scheme | VGGM-Finetune | | | | LDCNN |
|---|---|---|---|---|---|---|
| | | Fc1 | Fc1_ReLU | Fc2 | Fc2_ReLU | |
| **UCMD** | 0.375 | 0.349 | 0.334 | 0.329 | 0.34 | **0.439** |
| **RSD** | 0.279 | 0.171 | 0.062 | 0.028 | 0.022 | **0.019** |
| **RSSCN7** | 0.388 | 0.361 | 0.337 | 0.299 | 0.308 | **0.305** |

*4.4 Results of the Proposed LDCNN*

The proposed LDCNN result in a 30-D dimensional feature vector which is very compact compared to Fc and Conv features extracted by the pre-trained CNN. Table 8 shows the results of the proposed LDCNN on the three datasets. We can see LDCNN greatly improves the best results of the first scheme on RSD and RSSCN7 datasets by 26% and 8.3% respectively, and the performance is even slightly better than that of the fine-tuned Fc features. However, LDCNN does not achieve better performance than the first scheme and the fine-tuned VGGM model on UCMD dataset as expected. These results make sense because LDCNN is trained on AID dataset which is quite different from UCMD dataset but similar to the other two datasets and in particular RSD dataset in terms of image size and spatial resolution. Figure 11 shows some example images of the same class taken from the four datasets. It is evident that the images of UCMD are at a different zoom level with respect to the images of AID dataset, while the images of RSD and RSSCN7 (especially RSD) are similar to that of AID dataset. In addition, we can see the VGGM model fine-tuned on AID dataset only achieves slightly better performance than the first scheme on UCMD dataset, which also indicates that the worse performance of LDCNN is due to the difference between UCMD and AID datasets.

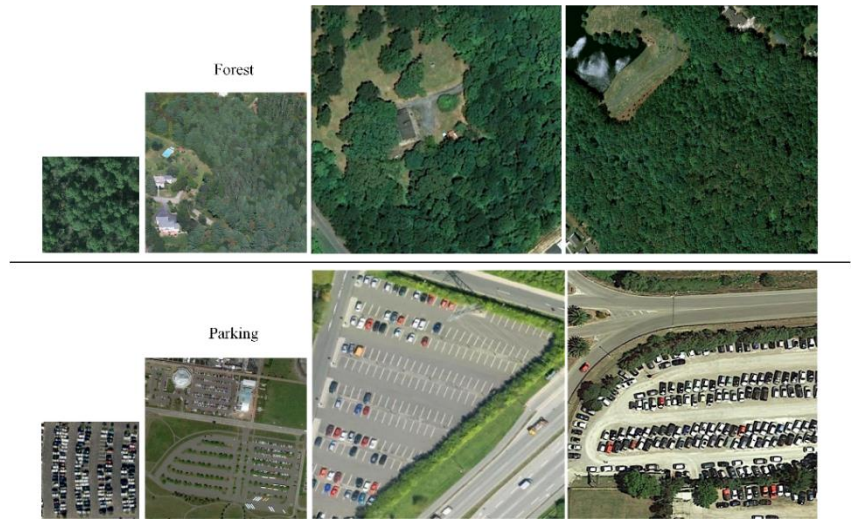

**Figure 11**. Comparison between images of the same class from the four datasets. From left to right, the images are from UCMD, RSSCN7, RSD and AID respectively.

Figure 12 shows the number of parameters (weights and biases) contained in the mlpconv layer of LDCNN and in the three Fc layers of VGGM and VGGM-Finetune. It is observed that LDCNN has about 2.6 times less parameters than VGGM-Finetune and 2.7 times less parameters than VGGM. Thus LDCNN is more easily to train if we use the same training dataset.

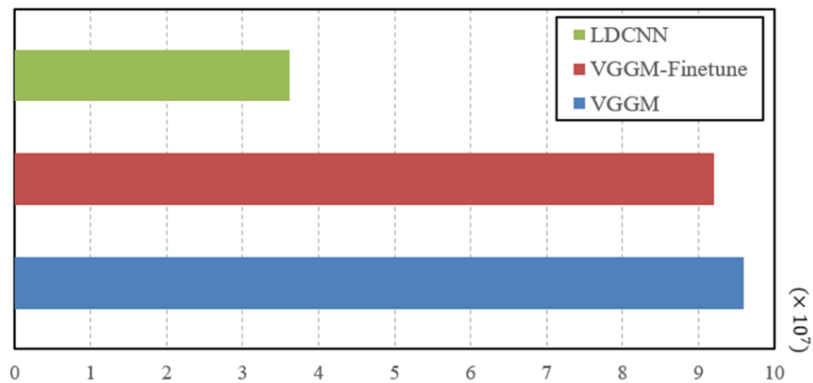

**Figure 12**. Comparison between parameters of the three models: VGGM, fine-tuned VGGM and LDCNN.

## 5. Discussion

From the extensive experiments, the two proposed schemes have been proven to be effective methods for HRRSIR. Some practical observations from the experiments are summarized in the following.

- The deep feature representations extracted by the pre-trained CNN and LDCNN achieve comparative or superior performance when compared to state-of-the-art hand-crafted features. The results indicate that CNN can generate powerful feature representations for HRRSIR task.
- In the first scheme, the pre-trained CNN is regarded as a feature extractor to extract Fc and Conv features. Fc features can be directly extracted from CNN's Fc1 and Fc2 layers, while feature aggregation methods (BOVW, VLAD and IFK) are needed in order to extract Conv features. To give an extensive evaluation of the deep features, we investigate the effect of ReLU on Fc and Conv features. The results show that Fc1 feature achieves better performance than Fc2 feature without the use of ReLU but worse performance with the use of ReLU, indicating nonlinearity will contribute to the improvement of retrieval performance if the features are extracted from higher layers. Regarding Conv features, it is interesting to find that the use of ReLU decreases the performance of Conv features on the four datasets except for UCMD dataset. This maybe because these existing feature aggregation methods are designed for traditional hand-crafted features which have quite different distributions of pairwise similarities from deep features [41].

- In the second scheme, a novel CNN that can learn low dimensional features for HRRSIR is proposed based on conventional convolution layer and a three-layer perceptron. In order to speed up training, the parameters of the convolutional layers are transferred from VGGM. The proposed LDCNN is able to generate 30-D dimensional feature vectors which are more compact than Fc and Conv features. As shown in Table 8, LDCNN outperforms the first scheme and the fine-tuned VGGM model on RSD and RSSCN7 datasets but achieves worse performance on UCMD dataset because the images in UCMD and AID are quite different in terms of image size and spatial resolution. LDCNN provides a direction for us to directly learn low dimensional features from CNN which can achieve remarkable performance.
- LDCNN is trained on AID dataset and then applied to three new remote sensing datasets (UCMD, RSD and RSSCN7). The remarkable performances on RSD and RSSCN7 datasets indicate that CNN has strong transferability. However, the performance should be further improved if LDCNN is trained on the target dataset. To this end, a much larger dataset such as the dataset used in Terrapattern project [42] needs to be constructed.

## 6. Conclusions

We present two effective schemes to extract deep feature representations for HRRSIR. In the first scheme, the features are extracted from the fully-connected and convolutional layers of a pre-trained CNN, respectively. The Fc features can be directly used for similarity measure, while the Conv features are encoded by feature aggregation methods to generate compact feature vectors before similarity measure. We also investigate the effect of ReLU on Fc and Conv features in order to achieve the best performance.

Though the first scheme can achieve better performance than traditional hand-crafted features, the pre-trained CNN models are trained on ImageNet which is quite different from remote sensing dataset. Thus we propose a novel CNN architecture based on conventional convolution layers and a three-layer perceptron which is then trained on a large remote sensing dataset. The proposed LDCNN is able to generate 30-D dimensional features that can achieve remarkable performance on several remote sensing datasets.

LDCNN is designed for HRRSIR, it can also be applied to other remote sensing tasks such as scene classification and object detection.

**Acknowledgments**

The authors would like to thank Paolo Napoletano for his code used for performance evaluation. This work was supported by the National Key Technologies Research and Development Program (2016YFB0502603), Fundamental Research Funds for the Central Universities (2042016kf0179 and 2042016kf1019), Wuhan Chen Guang Project (2016070204010114), Special task of technical innovation in Hubei Province (2016AAA018) and Natural Science Foundation of China (61671332).

**Author Contributions**

The research idea and design was conceived by Weixun Zhou and Zhenfeng Shao. The experiments were performed by Weixun Zhou and Congmin Li. The manuscript was written by Weixun Zhou. Shawn Newsam gives many suggestions and helps revise the manuscript.

**Conflicts of Interest**

The authors declare no conflict of interest.